\documentclass[letterpaper]{article} 
\usepackage[submission]{aaai23}  
\usepackage{times}  
\usepackage{helvet}  
\usepackage{courier}  
\usepackage[hyphens]{url}  
\usepackage{graphicx} 
\usepackage{natbib}  
\usepackage{caption} 
\usepackage{algorithm}
\usepackage{algorithmic}

\usepackage{newfloat}
\usepackage{listings}
\DeclareCaptionStyle{ruled}{labelfont=normalfont,labelsep=colon,strut=off} 
\floatstyle{ruled}
\newfloat{listing}{tb}{lst}{}
\floatname{listing}{Listing}
\title{OPHAvatars: One-shot Photo-realistic Head Avatars}
\author {
    Shaoxu Li,\textsuperscript{\rm 1}
}
\affiliations {
    \textsuperscript{\rm 1} John Hopcroft Center for Computer Science\\
    Shanghai Jiao Tong University\\
    Shanghai, China\\
    lishaoxu@sjtu.edu.cn
}
\usepackage{bibentry}

\begin{document}


\twocolumn[{%
\renewcommand\twocolumn[1][]{#1}%
\maketitle
\begin{center}
    \centering
    \captionsetup{type=figure}
    \includegraphics[width=1\textwidth]{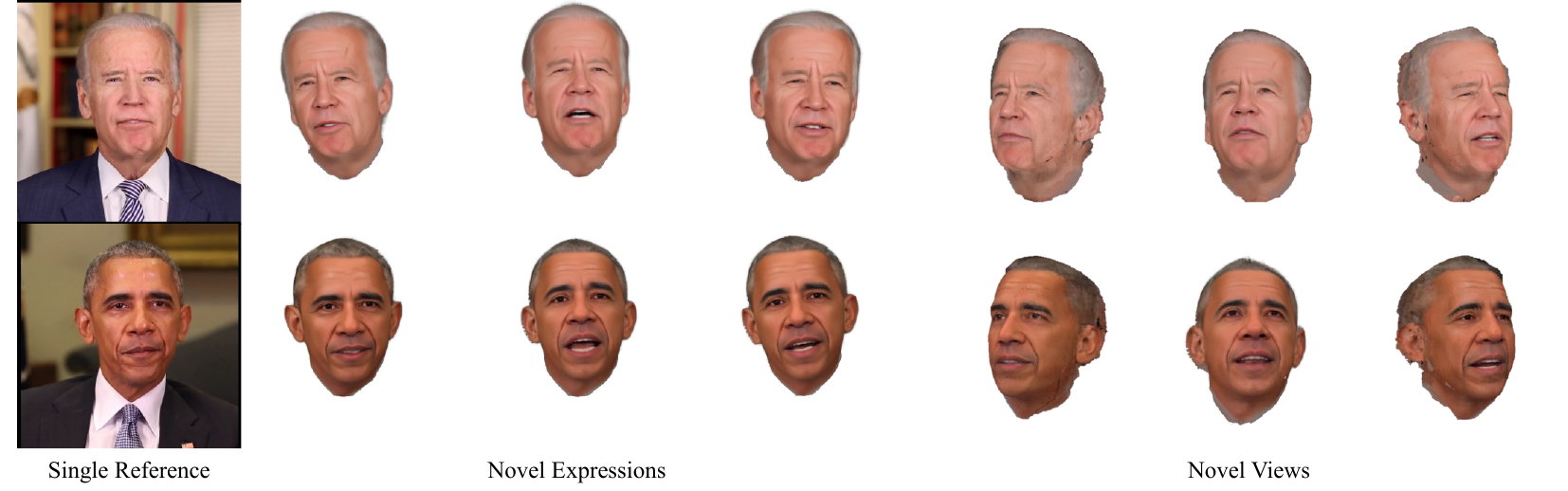}
    \captionof{figure}{Given a single reference image, our method optimizes a deformable neural radiance field to synthesize photo-realistic animatable 3D neural head avatars. The resulting head avatar can be viewed under novel views and animated with novel expressions.}
    \label{title}
\end{center}%
}]

\begin{abstract}
We propose a method for synthesizing photo-realistic digital avatars from only one portrait as the reference. Given a portrait, our method synthesizes a coarse talking head video using driving keypoints features. And with the coarse video, our method synthesizes a coarse talking head avatar with a deforming neural radiance field. With rendered images of the coarse avatar, our method updates the low-quality images with a blind face restoration model. With updated images, we retrain the avatar for higher quality. After several iterations, our method can synthesize a photo-realistic animatable 3D neural head avatar. The motivation of our method is deformable neural radiance field can eliminate the unnatural distortion caused by the image2video method. Our method outperforms state-of-the-art methods in quantitative and qualitative studies on various subjects.

\end{abstract}

\section{Introduction}

Recently, some methods have synthesized photo-realistic avatars leveraging neural radiance fields. As the most popular method for novel view synthesis, NeRF received much attention because of its superb photo-realistic rendering quality. Avatar creation and animation is a prevalent research task with broad application prospects. It's natural to explore the avatar generation with NeRF. Although some worthy works have been proposed, a one-shot avatar from an image is still underexplored. This work explores the photo-realistic avatar synthesis from a single portrait image.

Earlier works warp the source image with motion fields between the source and driving images. These methods try to simulate the 3D motion with extracted keypoints or other features. Some following works tend to resort to 3D head priors for better performance. Compared with 2D methods, 3D methods can better leverage the human face's unique geometry. In contrast to warp methods, generative methods also perform well, with adequate training images. For face rendering, explicit and implicit methods have strong points and weaknesses. 

In this paper, we propose one-shot photo-realistic head avatars(OHPAvatars), which can synthesize a high-quality animatable avatar with only a reference face image. Figure \ref{title} shows two animation examples. Given a single reference image, OHPAvatars can synthesize a high-quality avatar animated with novel expressions and views. Although plenty of works can accomplish dynamic avatar synthesis, the research of one-shot avatars from images remains underexplored. HeadNeRF\cite{hong2021headnerf} and OTAvatar\cite{ma2023otavatar} are similar works to ours. HeadNeRF can not preserve the identity well, and the dynamic rendering is unnatural. OTAvatar needs time-consuming training, and the rendering quality is unsatisfactory.

\begin{figure*}[ht]
\centering
\includegraphics[width=1\textwidth]{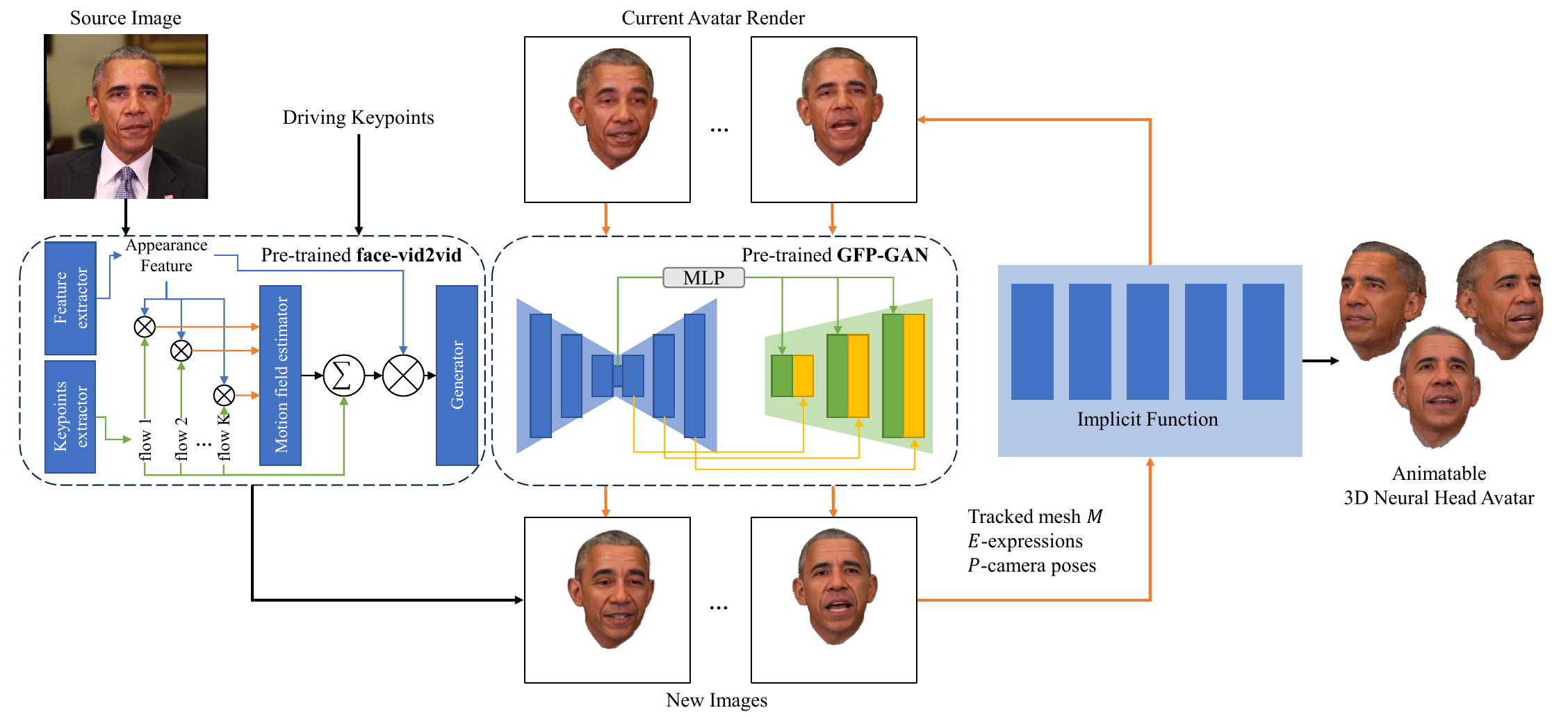} 
\caption{Overview. Our method synthesizes an animatable 3D neural head avatar and gradually updates the avatar by updating head images: (1) a coarse video is generated with a pre-trained face-vid2vid, given a source image and driving keypoints, (2) an animatable 3D neural head avatar is trained with the coarse video, (3) the avatar render images and update images with a pre-trained GFP-GAN, (4) animatable 3D neural head avatar continues training with new images. The iteration of (3)(4) repeats several times for high-quality avatar creation.}
\label{model}
\end{figure*}

Our method builds off video2avatar works with NeRF. Given a short video, a photo-realistic animated avatar can be obtained with a deformable neural radiance field, assisted by 3DMM priors. We employ an image2video method to get through the pipeline of image2avatar. To improve the poor rendering quality caused by inconsistency and low quality of the coarse video, we propose to update the avatar with a blind face restoration method iteratively. After the first avatar optimization with the coarse video, we render images with expressions and poses. We execute blind face restoration on all images and retrain the avatar. After several render-train, our method can synthesize a photo-realistic animatable 3D neural head avatar. Figure \ref{model} shows the detail of our pipeline.

In summary, our contributions are listed as follows:
\begin{itemize}
\item We propose a one-shot approach for creating a photo-realistic animatable 3D neural head avatar with a reference face image.

\item We propose eliminating the inconsistency of image-driven methods with neural radiance field and iteratively updating avatar quality with blind face restoration methods. The pipeline can be transferred to other domains.

\item We demonstrate remarkable results of our method through extensive experiments.

\end{itemize}

\section{Related Work}
\subsection{Talking Head}
Talking Head aims to synthesize photo-realistic videos, given source images and driving signals. Driving signals can be image, audio or expression coefficients. FOMM\cite{Siarohin_2019_fomm} proposed to extract keypoints mapping between source and driving images, warp the feature of the source image and then decoder for image synthesis. Face-vid2vid\cite{wang2021facevid2vid} proposed to estimate motion fields of feature points based on keypoints of source and driving images and warp the feature from source images based on the motion fields. Face2face\cite{Thies2016face2face} used 3D rendering to maintain the shape and illumination attributes when transferring expression and refining mouth details with a mouth retrieval algorithm. HeadGAN\cite{doukas2020headgan} used additional 3DMM mesh fitting results to assist the dense image flow prediction. DaGAN\cite{hong2022depth} proposed to capture dense 3D geometric information and estimate sparse facial keypoints for motion estimation. The depth information and motion fields warp the original image. StyleHEAT\cite{Yin2022StyleHEAT} integrated pre-trained StyleGAN\cite{Karras2020ada} to generate high-resolution talking face prediction by warping low-resolution feature maps. PIRenderer\cite{ren2021pirenderer} used a subset of 3DMM parameters as motion descriptors, a mapping network maps the input motion descriptors to latent space variables, a warping network generates a rough image, and an editing network generates the final image by editing the rough result.

Some methods pursued audio-driven talking head synthesis. Wav2Lip\cite{Prajwal2020Lip} constructed the encoder-decoder structure by using audio, pictures synchronized with audio and pictures not synchronized with audio as input, and realizes the audio-driven mouth video. MakeItTalk\cite{Yang2020MakeItTalk} used a pre-trained model to extract facial feature points and trained a baseline model of speech-driven feature points. SadTalker\cite{zhang2022sadtalker} modeled the expression and head pose separately in 3D face parameters and used the Wav2Lip model to complete the mouth synthesis.

\subsection{Neural Radiance Fields}
Neural radiance fields(NeRF)\cite{mildenhall2020nerf} is the latest method for novel view synthesis with differentiable volumetric rendering, which has received much attention because of its photo-realistic rendering quality. NeRF uses an MLP to store the 3D scene itself or its features. For avatar synthesis, characteristic features are embedded into the encoding parameters of NeRF. Some methods combine neural radiance fields and generative adversarial networks to complete scene synthesis and editing\cite{Schwarz2020GRAF, Niemeyer2021GIRAFFE, Eric2021piGAN, Chan2022Efficient}. Portrait manipulation can be accomplished with GAN inversion and latent code adjustment for these GAN-based methods. OTAvatar\cite{ma2023otavatar} proposed to decouple motion-related and motion-free latent code in inversion optimization by prompting the motion fraction of latent code ahead of the optimization using a decoupling-by-inverting strategy.

Others resort to 3DMM prior to realizing animated neural radiance field avatars. PVA\cite{raj2020pva} enabled portrait generation with a novel parameterization that combines NeRF with local, pixel-aligned features. AD-NeRF\cite{guo2021adnerf} was the first to propose a speech-driven neural radiance field head model, which completes the speech-to-speaker mapping. The method is trained on a speech video of the target person, and neural radiance fields implicitly represent the dynamic human head and upper body torso. HeadNeRF\cite{hong2021headnerf} collected and processed three large-scale face image datasets, based on which they designed a nerF-based parameterized head representation. NeRFace\cite{Gafni_2021_Dynamic} can synthesize high-quality faces based on controllable pose and expression, given a short face video. NeRFace extract pose and expression with 3DMM and encode them into the input of the neural radiance field. IMAvatar\cite{zheng2022imavatar} proposed to learn blendshapes and skinning fields to represent the expression-and pose-related deformations of implicit head avatars. NHA\cite{grassal2021neural} obtained high-precision meshes using a shape network and outputs high-quality synthetic images using a texture network. INSTA\cite{ZielonkaCVPR2023INSTA} adopted a similar strategy to IMAvatar, except that INSTA replaced blendshapes and skinning fields with mesh faces extracted by 3DMM and adopted multi-resolution hash coding to accelerate training.

\begin{figure*}[ht]
\centering
\includegraphics[width=0.95\textwidth]{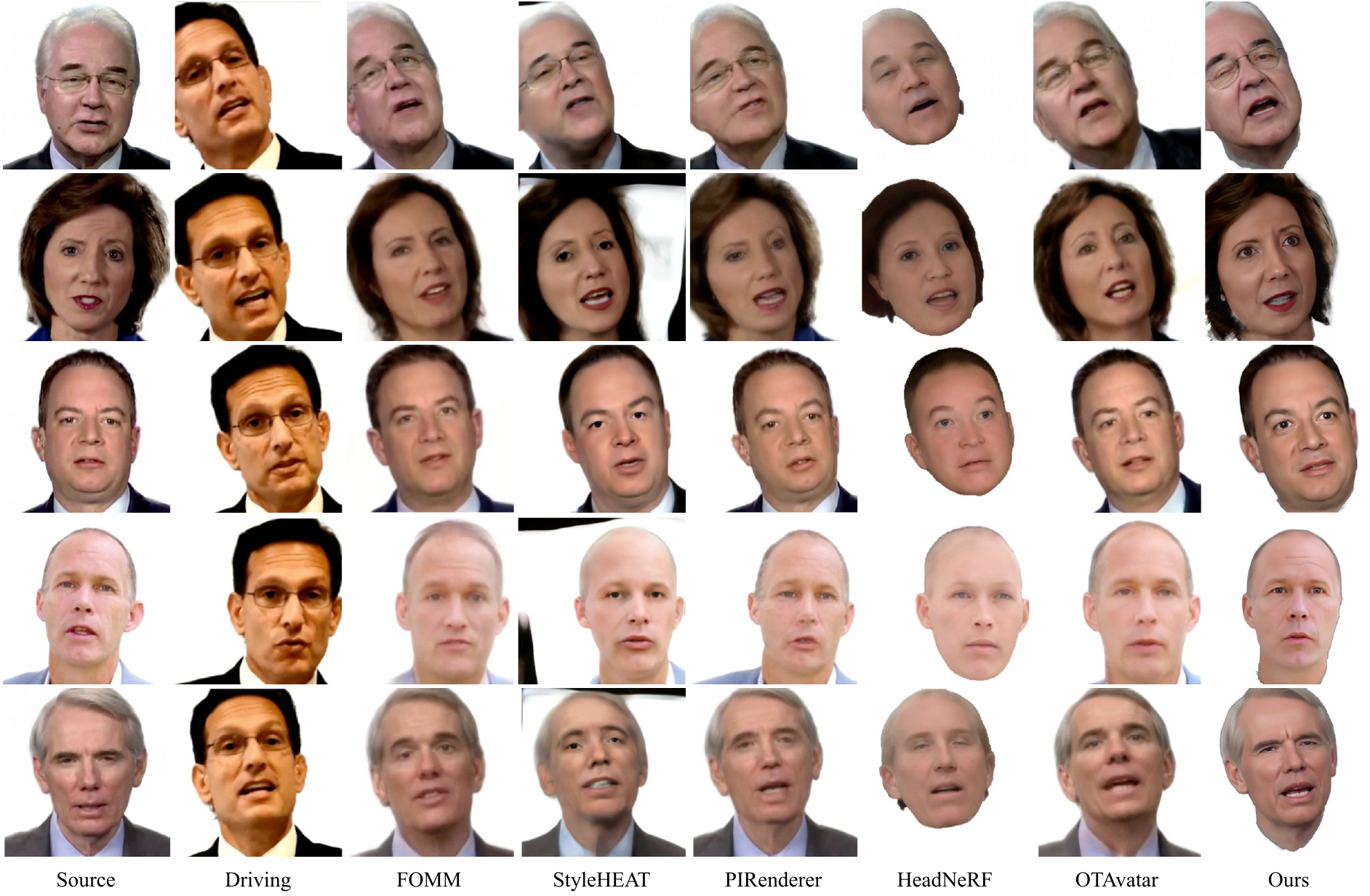} 
\caption{Qualitative result for cross-identity reenactment.}
\label{exp1}
\end{figure*}

\subsection{Blind Face Restoration}
Blind face restoration aims at recovering high-quality portraits from low-quality face photos. The restoration tasks include face denoising, face deblurring, face super-resolution, face artifact removal and blind face restoration. The restoration solutions include geometric prior-based methods, reference prior-based methods, generative prior-based methods and non-prior-based methods\cite{wang2022survey}. Geometric prior-based restoration methods utilize the unique facial geometry characteristics to restore face images, such as facial landmarks\cite{CT-FSRNet-2018,kim2019progressive}, face parsing maps\cite{ChenPSFRGAN} and facial heatmaps\cite{Yu2018Face}. Reference prior-based restoration methods use guidance from additional high-quality face images, such as the facial structure or facial component dictionaries\cite{Li_2018_Learning, Li_2020_Enhanced, Li_2020_Blind, Dogan_2019_Exemplar}. Generative models aim to synthesize high-quality images with abundant training images from specific domains. Generative prior-based restoration methods leverage the generative capacity of generative models. For example, PLUSE\cite{menon2020pulse}, mGANprior\cite{gu2020image} restore face by optimizing the latent code from a pre-trained GAN. GFP-GAN\cite{wang2021gfpgan} and GPEN\cite{Yang2021GPEN} use pre-trained GAN as a decoder for face restoration.

Non-prior-based restoration methods try to establish the mapping between low-quality and high-quality face images without any facial priors. For example, BCCNN\cite{Zhou2015Learning} extracts robust face representations from the low-resolution face image and fuses the representation and the input image for restoration. CBN\cite{zhu2016deep} jointly optimizes FSR and dense correspondence field estimation in a deep bi-network, which benefit each other and recover high-quality face images.

Recently, some methods leveraged attention mechanisms to extract face features, which improve the performance\cite{ChenSPARNet}. And some methods build off the transformer backbone to build the networks more recently\cite{wang2022restoreformer,zhang2022blind}.

\section{Method}
Our method takes as input a human portrait image. As output, our method produces a photo-realistic animatable 3D neural head avatar.

Our method synthesizes a coarse talking head video using driving keypoints features. With the video, we extract parameters and construct a neural radiance field in the canonical space to represent the head and deform the radiance field to learn the avatar with facial expressions. For a photo-realistic avatar, we execute blind face restoration with rendered face images and iteratively update the face images for more iterations.

Our method builds off three parts, talking head from one portrait image, blind face restoration, photo-realistic 3D neural head avatar from a video, specifically face-vid2vid\cite{wang2021facevid2vid}, GFPGAN\cite{wang2021gfpgan} and INSTA\cite{ZielonkaCVPR2023INSTA}.

\subsection{Background}
\subsubsection{Face-vid2vid.}
Face-vid2vid\cite{wang2021facevid2vid} extracts appearance features $F_{appearance}$, 3D canonical keypoints ${x_{c,k}}$, head pose $R_s,t_s$ and the expression deformations $\delta_{s,k}$ from the source image. Only driving head pose $R_d,t_d$ and the expression deformations $\delta_{d,k}$ are extracted from the driving images. Source keypoints ${x_{s,k}}$ are calculated with ${x_{c,k}}$, $R_S,t_S$ and $\delta_{S,K}$. Driving keypoints ${x_{d,k}}$ are calculated with ${x_{c,k}}$, $R_d,t_d$ and $\delta_{d,k}$. Face-vid2vid estimate motion flows $f_s$ from the source and driving keypoints for video synthesis. The flow results are combined and fed to the motion field estimation network to produce a flow composition mask $m$. A linear combination of $m$ and motion flows ${f_s}$ produces the composited flow field $w$, which warp the appearance features $F_{appearance}$. Finally, a generator converts the warped feature $w(F_{appearance})$ to the output image. For more details, we direct the reader to the original paper\cite{wang2021facevid2vid}.

\subsubsection{GFPGAN.}
GFPGAN\cite{wang2021gfpgan} removes the image degradation and extracts features $F_{latent}$ and $F_{spatial}$ with a degradation removal module, which is a U-Net structure. 
\begin{equation}
    F_{latent},F_{spatial}=\rm{U \mbox{-} Net}(x),
\end{equation}
GFPGAN uses the latent features $F_{latent}$ to map the input image to the closest latent codes $W$ in StyleGAN2\cite{}. The latent codes $W$ generate the GAN features $F_{GAN}$.
\begin{equation}
    W ={\rm{MLP}}(F_{latent}),
\end{equation}
\begin{equation}
    F_{GAN}={\rm{StyleGAN}}(W),
\end{equation}
GFPGAN uses the multi-resolution spatial features $F_{spatial}$ to modulate the StyleGAN2 features. At each resolution scale, a pair of affine transformation parameters $\alpha,\beta$ are generated from spatial features $F_{spatial}$. And the modulation is carried out through channel-split spatial feature transform (CSSFT) layers. For more details, we direct the reader to the original paper\cite{wang2021gfpgan}.
\begin{equation}
    \alpha,\beta ={\rm{Conv}}(F_{spatial}),
\end{equation}
\begin{equation}
    F_{output}={\rm{SC\mbox{-}SFT}} (F_{GAN}|\alpha,\beta),
\end{equation}

\begin{figure*}[ht]
\centering
\includegraphics[width=0.95\textwidth]{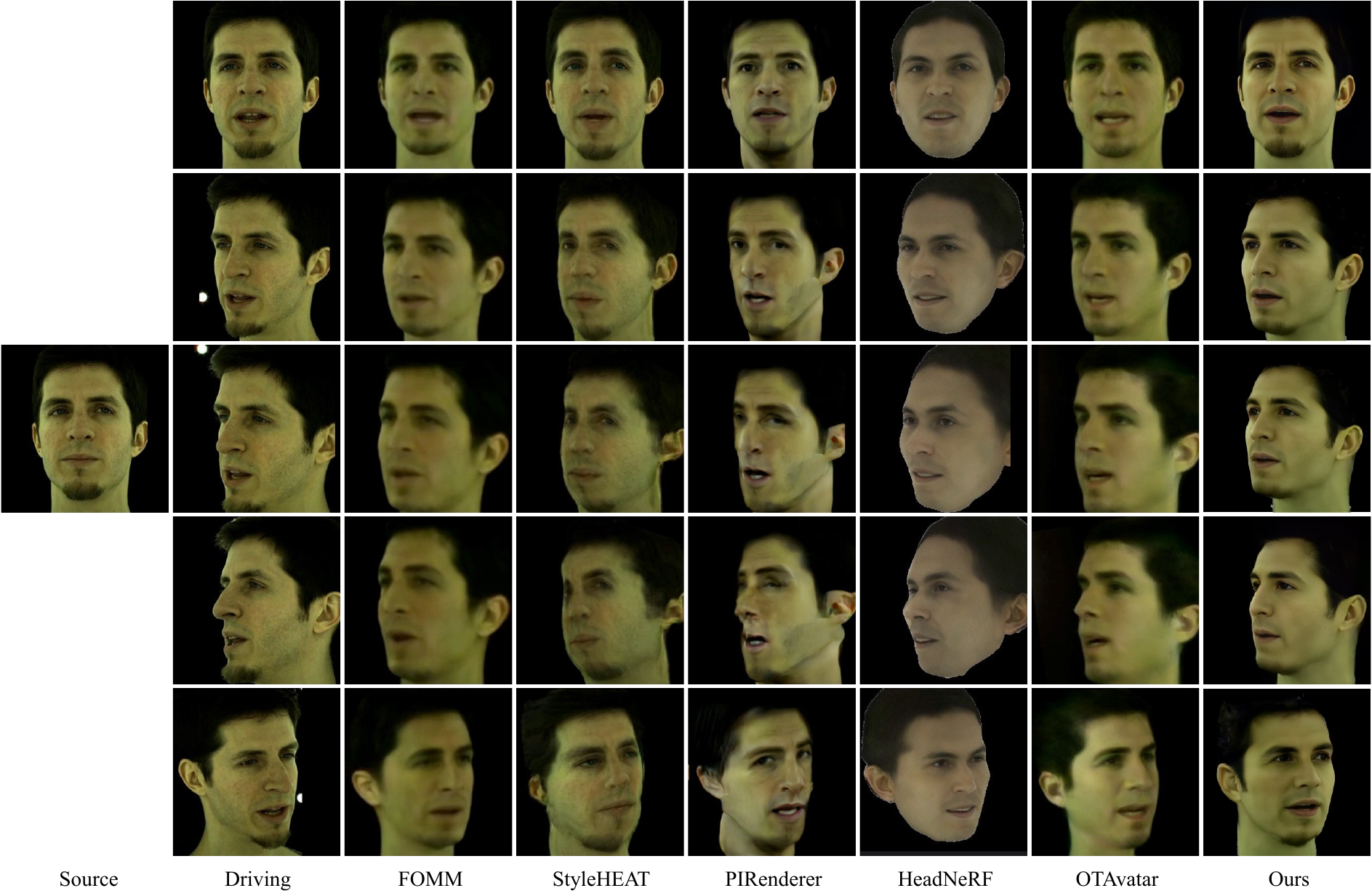} 
\caption{Qualitative result for multi-view reenactment.}
\label{exp2}
\end{figure*}

\subsubsection{INSTA.}
Neural radiance field\cite{mildenhall2020nerf} based head synthesis has attracted much attention for its high quality. INSTA\cite{ZielonkaCVPR2023INSTA} is a neural radiance field-based method for photo-realistic digital avatars reconstructing. In practice, INSTA is trained on a single monocular RGB portrait video and can reconstruct a digital avatar that extrapolates to unseen expressions and poses. For a given video, inputs for avatar optimization include head images ${I_i}$, the intrinsic camera parameters $K\in{R^{3\times 3}}$, tracked FLAME\cite{li2017flame} meshes $\{M_i\}$, facial expression coefficient $\{E_i\}$, and head poses $\{P_i\}$. INSTA constructs a neural radiance field in the canonical space to represent the head. A mapping function $\phi(p, M_i)$ projects the points from time-varying deformed space to the canonical space, leveraging the tracked FLAME\cite{li2017flame} meshes $\{M_i\}$ and a predefined mesh in canonical space $\{M_canon\}$. Differentiable volumetric rendering is used to optimize the radiance field. The animatable dynamic neural radiance field is implemented using the Nvidia NGP C++ framework\cite{}. In practice, INSTA samples 1700 frames from the video and trains the network for 32k optimization steps. The resulting head avatar can be viewed under novel views and expressions with the optimized deformable radiance field. For more details, we direct the reader to the original paper\cite{ZielonkaCVPR2023INSTA}.

\subsection{One-shot Photo-realistic Head Avatars}
A short monocular RGB video can reconstruct neural radiance field-based photo-realistic digital avatars. Our method works by generating a coarse video with a pre-trained image2video model and embedding the data enhancement into the training of the neural radiance field by updating the rendered images with a blind face restoration model.

In this section, we first illustrate the process of generating a coarse video, then describe the iterative dataset update strategy, and finally, discuss our method's implementation details. The following mentioned head images are segmented from original videos, which do not include background and body.

\subsubsection{Coarse Avatar Generation.}
We use face-vid2vid\cite{wang2021facevid2vid} to obtain a coarse video. As a talking head synthesis method, it inputs a source image $I_s$ and a sequence of unsupervised-learned 3D keypoints ${x_{d,k}}$. The output coarse talking head video contains a set of portrait images with different expressions and poses, which provide abundant training information for avatar synthesis. With these initial portrait images, we segment the background and foreground to obtain training images $\{I_{d, i}^{coarse}\}$. FLAME meshes and parameters are calculated for later avatar optimization. We use INSTA\cite{ZielonkaCVPR2023INSTA} to train an animatable 3D neural head avatar, a deformable neural radiance field with 3DMMs parameters as reference.

With the increase of view angle, the synthesis quality of the coarse video decreases. For avatar synthesis, large view angle images can improve the quality of the avatar. Fortunately, as a 3D representation, NeRF can maintain the stability of the scene geometry. In this case, the stable facial geometry promises the stability of the identity. For the deformable neural radiance field, the coarse video's distortions only impact the avatar's visual quality but not the geometry. But we cannot obtain high-quality avatars with low-quality images from the coarse video.

\begin{figure*}[ht]
\centering
\includegraphics[width=0.95\textwidth]{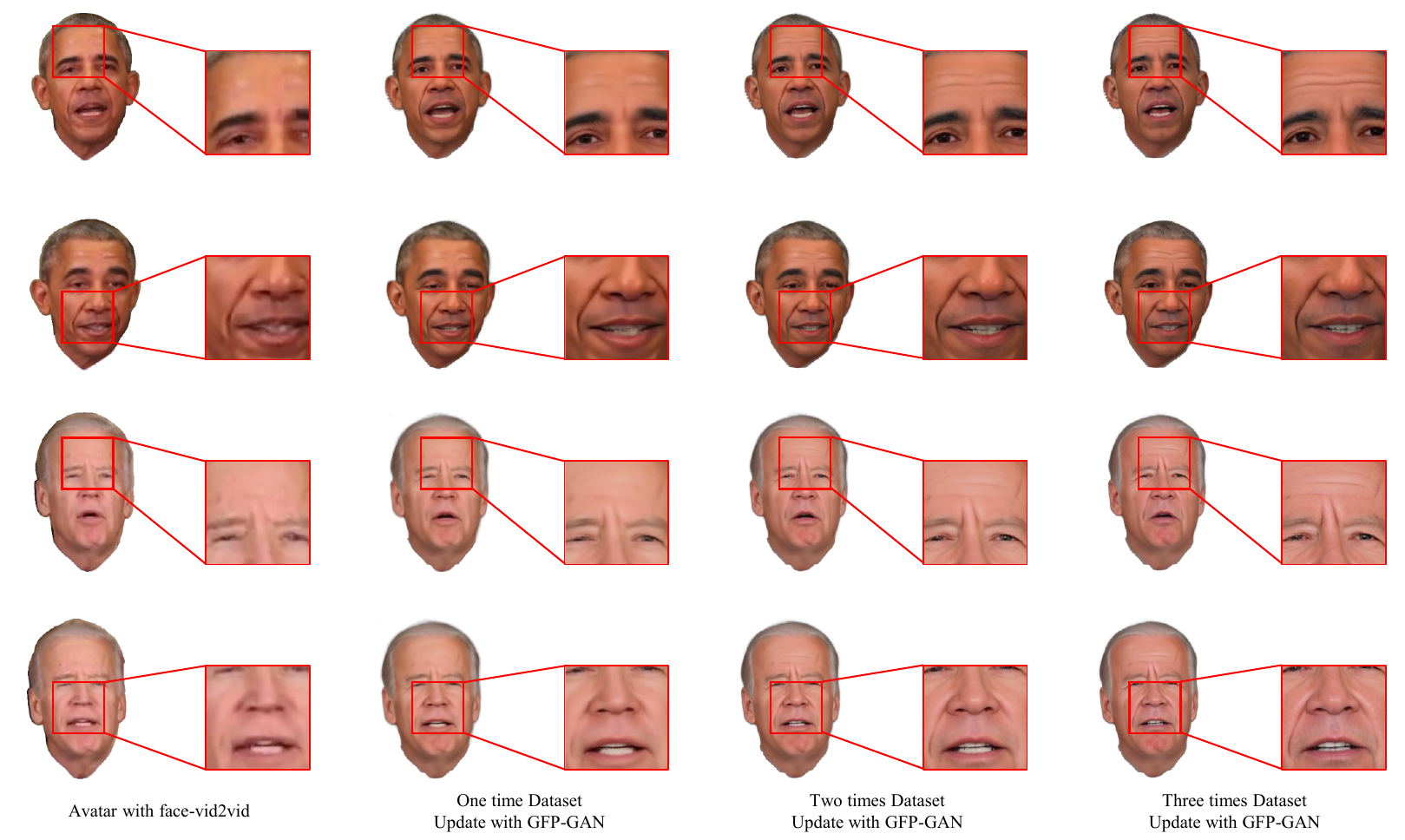} 
\caption{Ablation Study: We compare face-vid2vid and one/twp/three times dataset update avatar synthesis results with GFP-GAN.}
\label{ablation}
\end{figure*}

\subsubsection{Iterative Dataset Update.}
Even though NeRF is a 3D representation method that ensures high-quality image synthesis, our avatar is of low quality due to the initial video's poor image quality and inconsistency. For high-quality synthesis, we propose an iterative dataset update with blind face restoration. After acquiring the coarse avatar, we render the avatar into images $\{I_{d, i}^{render}\}$, conditioned with various expressions and views. Then a face image restoration is executed on all the images, specifically GFPGAN\cite{wang2021gfpgan}. With new images and rendered parameters, we train the avatar again. The process above can be conducted repeatedly. The rendering quality of the avatar increase with the iteration. After several iterations, we can obtain a photo-realistic animatable 3D neural head avatar.

\begin{figure}[H]
    \centering
    \includegraphics[width=1\linewidth]{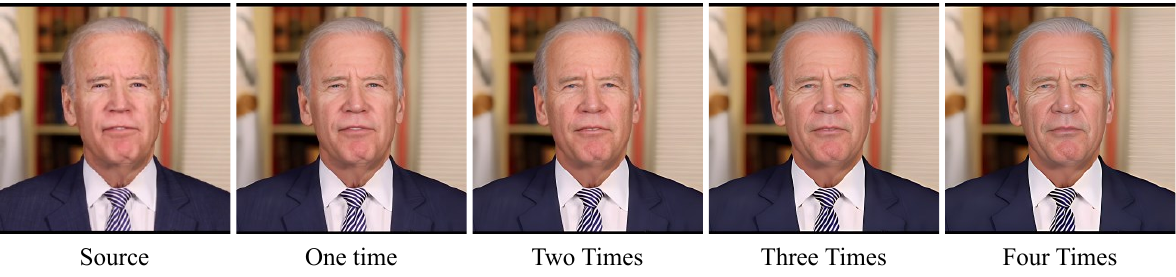}
    \caption{The face restoration results with the iterations.}
    \label{fig:gfpgan}
\end{figure}

Although the rendering quality increases with iteration, the results may not be as expected. For a coarse image, a restoration tries to fill in details. A well-restored image may be inconsistent with the original image, although clear in detail. Figure\ref{fig:gfpgan} shows the restoration results of a portrait image. The image becomes sharper with the iterations, but too many details are added.

\begin{table*}[htbp]
  \centering
  \caption{Quantitative comparisons on multi-View reenactment and cross-identity reenactment.}
    \scalebox{0.9}{\begin{tabular}{c|cccccccc|ccccc}
          & \multicolumn{8}{c|}{Multi-View Reenactment}                   & \multicolumn{5}{c}{Cross-Identity Reenactment} \\
    \hline \rule{0pt}{8pt}
    & PSNR$\uparrow$  & SSIM$\uparrow$  & CSIM$\uparrow$  & AED$\downarrow$   & APD$\downarrow$   & AKD$\downarrow$   & LIPIPS$\downarrow$ & FID$\downarrow$   & CSIM$\uparrow$  & AED$\downarrow$   & APD$\downarrow$   & AKD$\downarrow$   & FID$\downarrow$ \\
    \hline
    FOMM  & 20.75  & 0.639 & 0.505 & 2.004 & 0.545  & 5.052 & 0.308 & 101.6 & 0.672 & 3.196  & 0.500  & 4.198 & 113.2  \\
    PIRenderer & 20.04  & 0.586 & 0.493 & 2.203 & 0.680  & 6.566 & 0.299 & 100.6 & 0.632 & 3.018  & 0.498  & 4.977 & 103.7  \\
    StyleHEAT & 20.03  & 0.632 & 0.387 & 2.179 & 0.472  & 5.522 & 0.284 & 123.8 & 0.614 & 2.860  & 0.471  & 3.592 & 239.1  \\
    HeadNeRF & 17.60  & 0.546 & 0.239 & 2.086 & 0.776  & 4.166 & 0.367 & 212.3 & 0.282 & 2.873  & 0.567  & \textbf{3.465} & 233.0  \\
    OTAvatar & 21.19  & 0.657 & 0.574 & 1.874 & 0.428  & \textbf{3.731} & 0.288 & 137.3 & \textbf{0.694} & 2.850  & 0.405  & 4.307 & 101.8  \\
    Ours  & \textbf{21.68 } & \textbf{0.690 } & \textbf{0.634} & \textbf{1.495} & \textbf{0.265} & 4.135  & \textbf{0.231} & \textbf{95.3} & 0.503 & \textbf{1.439} & \textbf{0.270 } & 4.245 & \textbf{97.4} \\
    \hline
    \end{tabular}%
    }
  \label{tab:quantitative}%
\end{table*}%

\section{Results}
\subsection{Experiment Setting}
In this section, we first describe the implementation details of our method. Then we compare our method with state-of-art face reenactment and avatar creation methods. At last, we execute an ablation study to illustrate the effectiveness of our designs.

\subsubsection{Implementation details.}
Abundant poses and expressions are needed for high-quality avatar training. We create around 2-3min videos with manually driving keypoints or driving videos for coarse video generation. The coarse video is sub-sampled to 25fps and resized to 512×512 resolution. We preprocess the video with the pipeline in INSTA\cite{ZielonkaCVPR2023INSTA}. Firstly, we use background foreground segmentation using robust matting\cite{Shanchuan2021Robust} and an off-the-shelf face parsing framework\cite{yu2020bisenet} for image segmentation and clothes removal. Secondly, we use MICA\cite{Zielonka2022Towards} to obtain optimized FLAME mesh, expression parameters, and pinhole camera intrinsic and extrinsic parameters. With these parameters and head images, we use INSTA to optimize an avatar. Our model is implemented in Pytorch using one RTX 3090. For each avatar optimization, the model is trained for 30000 iterations. Thanks to the fast training of INSTA with multiresolution encoding, each iteration can be accomplished in less than 20 minutes.

\subsubsection{Dataset.}
We conduct cross-identity reenactment experiments on HDTF\cite{zhang2021flow} dataset and multi-view reenactment experiments on Multiface\cite{wuu2023multiface} dataset. And the quantitative comparison is conducted on the same dataset. For the ablation study, we use two celebrity photos for illustration.

\subsubsection{Baseline methods.}
We compare our method with state-of-the-art 2D and 3D methods. For image-driven method, FOMM\cite{Siarohin_2019_fomm} and face-vid2vid\cite{wang2021facevid2vid} achieve SOTA performance. For 2D methods with 3DMM coefficients, StyleHEAT\cite{Yin2022StyleHEAT} and PIRenderer\cite{ren2021pirenderer} are the SOTA methods. For 3D methods, HeadNeRF\cite{hong2021headnerf} and OTAvatar\cite{ma2023otavatar} achieve SOTA perfomance. As a component of our method, face-vid2vid\cite{wang2021facevid2vid} is compared in the ablation study.

\subsubsection{Evaluation metrics.}
For comprehensive comparative analysis, we adopt peak-signal-to-noise ratio (PSNR), structural similarity index measure (SSIM), and learned perceptual image patch similarity (LPIPS)\cite{zhang2018perceptual} for visual quality, frechet inception distance (FID)\cite{Heusel2017GANs} and cosine similarity of identity embedding (CSIM)\cite{Deng_2022ArcFace} for image realism. And inspired by OTAvatar\cite{ma2023otavatar}, we adopt average expression distance (AED), average pose distance (APD), and average keypoint distance (AKD) for facial expression and pose transfer.

\subsection{Evaluation Result}
\subsubsection{Qualitative comparison.}
For cross-identity reenactment, we evaluate the transfer quality of the synthesis results on HDTF\cite{zhang2021flow} dataset. Figure \ref{exp1} shows the qualitative results. Compared with warping-based methods FOMM, StyleHEAT and PIRender, our methods can maintain the identity better and produce consistent facial geometry. Compared with 3D HeadNeRF, our method reconstruct the identity of the source image better. Compared with OTAvatar, our method ensures higher visual quality. Some pose and expression biases are caused by inaccurate poses and expressions extraction from driving images. And some facial details vary from the source images since our avatar is optimized with the blind face restoration method.

For multi-view reenactment, we evaluate the consistency of the synthesis results in different views on Multiface\cite{wuu2023multiface} dataset. Figure \ref{exp2} shows the qualitative results. Compared with cross-identity reenactment, multi-view reenactment shows each method's large view angel face reenactment ability. StyleHEAT and PIRender can hardly maintain the face geometry. The face identity of HeadNeRF is far from the source image. FOMM and OTAvatar can maintain the face geometry well, while the rendering quality is unsatisfactory. By our method, the avatar can preserve the identity well, with the best rendering quality.

\subsubsection{Quantitative comparison.}
Table \ref{tab:quantitative} shows the quantitative comparison results on multi-view reenactment and cross-identity reenactment. Multi-View Reenactment is computed on Multifaces datasets, and Cross-Identity Reenactment is computed on HDTF datasets. Our method outperforms other methods in terms of most of the criteria.

\subsection{Ablation Study}
We validate the architecture of our method by ablation study. Figure \ref{ablation} shows the avatar animation results with coarse video from face-vid2vid and one/two/three times dataset update with GFP-GAN. The two celebrity photos in Figure \ref{title} are the source images. The driving poses are extracted from a short video. The dataset update makes rendering images sharper, and the rough facial structures remain unchanged. But the sharp image may be inconsistent with the source image. We suggest using dataset update two times, weighing the source identity and the avatar quality.

\subsection{Limitations}
Our method uses blind face restoration to improve the avatar's quality, which leads to many facial details. Wrinkles from the original portrait are sharpened. More details may deviate from the original identity. Although our method can synthesize novel view images for unseen poses, it's hard to explore extreme novel views, for example, 90◦. With the increase of view angle, the rendering quality decreases. With the increase of view angles, the expression coefficients extracted may be inaccurate, which leads to inaccurate rendering expressions.

\section{Conclusion}
In this work, we present a novel framework of one-shot photo-realistic head avatars, namely OPHAvatars, using a deformable neural radiance field. The pipeline includes two stages, coarse video generation and neural avatar synthesis. Coarse video generation synthesizes a talking head video with pre-trained image2video methods. Neural avatar synthesis takes the coarse video as input and synthesizes an animatable avatar with a deformable neural radiance field. Then we take a dataset update strategy to improve the performance with a blind face restoration method. After several restoration and training, our method optimizes a photo-realistic animatable 3D neural head avatar. Comprehensive experiments were conducted to prove the superiority of our proposed framework.

\bibliography{aaai23}

\end{document}